\documentclass{preprint}

\usepackage{lipsum} 
\usepackage{xcolor}
\usepackage{booktabs}
\usepackage{graphicx} 
\usepackage[numbers,comma,sort&compress,super]{natbib}
\usepackage[colorlinks=true]{hyperref}
\usepackage{subcaption}
\usepackage{tabularx}
\usepackage{colortbl}
\usepackage{pdflscape}
\usepackage{multirow}
\usepackage{makecell}

\title{Machine Learning Applications Related to Suicide in Military and Veterans: A Scoping Literature Review}
\author[1,$\dagger$]{Yuhan Zhang}
\author[1,$\dagger$]{Yishu Wei}
\author[2,3]{Yanshan Wang}
\author[1]{Yunyu Xiao}
\author[4]{COL (Ret.) Ronald K. Poropatich}
\author[5,6]{Gretchen L. Haas}
\author[1]{Yiye Zhang}
\author[7]{Chunhua Weng}
\author[8]{Jinze Liu}
\author[9,10]{Lisa A. Brenner}
\author[11]{James M. Bjork}
\author[1,*]{Yifan Peng}
\affil[1]{Department of Population Health Sciences, Weill Cornell Medicine, New York, USA}
\affil[2]{Department of Health Information Management, University of Pittsburgh, Pittsburgh, USA}
\affil[3]{VA Pittsburgh Healthcare System, Pittsburgh, PA}
\affil[4]{Center for Military Medicine Research, University of Pittsburgh, Pittsburgh, PA}
\affil[5]{Department of Psychiatry, University of Pittsburgh Medical Center, Pittsburgh, USA}
\affil[6]{VISN 4 MIRECC, VA Pittsburgh Healthcare System, Pittsburgh, USA}
\affil[7]{Department of Biomedical Informatics, Columbia University, New York, USA}
\affil[8]{Department of Biostatistics, Virginia Commonwealth University, Richmond, USA}
\affil[9]{Department of Physical Medicine and Rehabilitation, University of Colorado, Anschutz Medical Campus, Aurora, USA}
\affil[10]{VA Brain Health Coordinating Center, Department of Veterans Affairs, Aurora, Colorado}
\affil[11]{VISN 6 MIRECC, Central Virginia VA Health Care System, Richmond, VA}
\affil[$\dagger$]{These authors contributed equally to this work.}
\affil[*]{Corresponding author(s). Email(s): \url{yip4002@med.cornell.edu}}

\begin{document}

\maketitle

\begin{abstract}
\textbf{Background:} Suicide remains one of the main preventable causes of death among active service members and veterans. Early detection and prediction are crucial in suicide prevention. Machine learning techniques have been explored in recent years with a specific focus on the assessment and prediction of multiple suicide-related outcomes, which have yielded promising results. This study aims to assess and summarize current research and provides a comprehensive review regarding the application of machine learning techniques in assessing and predicting suicidal ideation, attempts, and mortality among members of military and veteran populations.

\textbf{Methods:} A keyword search using PubMed, IEEE, ACM, and Google Scholar was conducted, and the PRISMA protocol was adopted for relevant study selection. Peer-reviewed original research in English targeting the assessment or prediction of suicide-related outcomes among service members and veteran populations was included. 1,110 studies were retrieved, and 32 satisfied the inclusion criteria and were included. 

\textbf{Outcomes:} Thirty-two articles met the inclusion criteria. Despite these studies exhibiting significant variability in sample characteristics, data modalities, specific suicide-related outcomes, and the machine learning technologies employed, they consistently identified risk factors relevant to mental health issues such as depression, post-traumatic stress disorder (PTSD), suicidal ideation, prior attempts, physical health problems, and demographic characteristics.
Machine learning models applied in this area have demonstrated reasonable predictive accuracy and have verified, on a large scale, risk factors previously detected by more manual analytic methods. Additional research gaps still exist. First, many studies have overlooked metrics that distinguish between false positives and negatives, such as positive predictive value and negative predictive value, which are crucial in the context of suicide prevention policies. Second, more dedicated approaches to handling survival and longitudinal data should be explored. Lastly, most studies focused on machine learning methods, with limited discussion of their connection to clinical rationales.

\textbf{Conclusion:} 
In summary, machine learning analyses have identified risk factors associated with suicide in military populations, which span a wide range of psychological, biological, and sociocultural factors, highlighting the complexities involved in assessing suicide risk among service members and veterans. Some differences were noted between males and females. The diversity of these factors also demonstrates that effective prevention strategies must be comprehensive and flexible. 
\end{abstract}


\section{Introduction}\label{introduction}

Preventing suicide poses a multifaceted challenge,  with particularly and persistently high rates noted among military personnel and veterans.\cite{hoge2012preventing-n, kuehn2009soldier-g, nock2014prevalence-a, ursano2015suicide-q}
Over the past two decades, the suicide rates among United States (U.S.) service members have risen significantly, for the first time exceeding the rates observed within the civilian population.\cite{rozek2020using-r}
Furthermore, statistics indicate that suicides now account for more deaths among U.S. military personnel than combat-related fatalities.\cite{colic2018using-z}
Mortality rates from suicide are also high among members of the veteran population. According to the 2024 National Suicide Prevention Annual Report, the age-adjusted suicide rate in 2022 among female veterans was 92.4\% higher than for non-veteran female U.S. adults, and this rate for male veterans was 44.3\% higher compared to non-veteran male U.S. adults.\cite{VA_Suicide_Prevention_Office_of_Mental_Health_and_Suicide_Prevention2023-ru}
Therefore, there is a critical need to review the relevant studies focused on identifying, addressing, and attempting to mitigate factors contributing to the heightened risk of suicide among members of both active duty and veteran communities.

Various prevention programs have been implemented to address the increasing issue of suicide within military and veteran populations. These initiatives include universal interventions applied to the entire demographic (e.g., Ask/Care/Escort model\cite{blocker2013unintended-t}), as well as the more focused interventions targeting individuals identified as high risk (e.g., post-deployment screening\cite{Panaite2018-oy}). 

Despite these efforts, data still show a persistent rise in suicide rates,\cite{thompson2014predicting-k} highlighting the need for greater community awareness, advanced methods for identifying risk factors, and more effective mitigation strategies. 
Data-driven approaches like machine learning (ML) have the potential to more quickly and effectively identify high-risk populations and individuals to enhance prevention strategies. ML algorithms are able to interpret large amounts of data efficiently, enabling the identification of effective factors and complex patterns for predicting suicide risk.\cite{colic2018using-z}
This capability has led to significant advancements in identifying and predicting suicide-related outcomes within military and veteran populations.\cite{bryan2021improving-t, vogt2024u-x, kessler2017predicting-s}
Importantly, ML-based detection of suicide risk factors in large health systems could be highly impactful, as these systems can apply the same variables to flag new patients or encounters using readily available data. 
For example, the VA ``REACH-VET'' program leverages previous findings from ML analysis of VA electronic medical records (EMR) to detect and flag those Veterans with ML-detected risk factors for clinician scrutiny.\cite{mccarthy2021evaluation-v, kessler2017developing-c, matarazzo2023veterans-d}

However, existing ML research remains fragmented, and a comprehensive review is needed to synthesize these findings and chart future research directions. To our knowledge, there are no comprehensive reviews synthesizing these machine-learning applications in studying active duty and veteran communities.\cite{naifeh2019army-y}
To address this gap, this study aims to conduct a scoping literature review to examine the application of ML in suicide risk assessment and prediction among active military and veteran populations. The goal is to systematically collect and summarize original research on this topic, emphasizing the potential of ML in addressing suicide within these groups. The study highlights the availability of comprehensive datasets for predicting suicide-related outcomes and identifies consistent risk factors found through machine learning despite methodological differences. Additionally, it seeks to pinpoint existing gaps in data practices and machine learning techniques and calls for collaboration among behavioral health professionals, community organizations, and ML researchers to mitigate suicide rates in this critical area.

\section{Methods}\label{methods}

\begin{figure}
    \centering
    \includegraphics[width=0.667\linewidth]{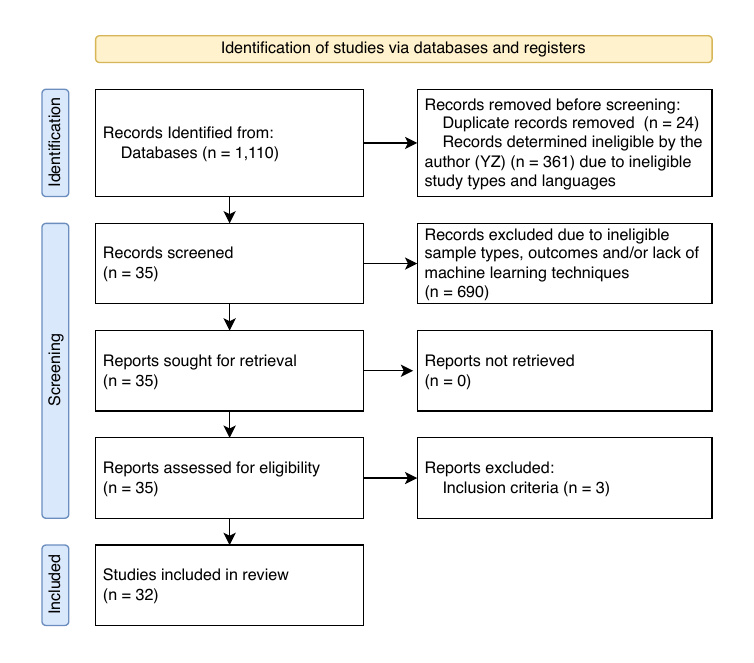}
    \caption{Systematic Reviews and Meta-Analyses (PRISMA) flow diagram.}
    \label{fig:flow}
\end{figure}

The PRISMA protocol was adopted to systematically review studies focused on the application of ML in suicide risk prediction among veterans. The process involved several stages, including defining eligibility criteria, sourcing information from various databases, employing a keyword search strategy, selecting studies based on specific inclusion criteria, and performing data analysis. Each stage is outlined below to provide a clear understanding of the methodology employed (Figure \ref{fig:flow}).

\subsection{Eligibility criteria}\label{eligibility-criteria}

The abstracts of the documents were assessed using the following eligibility criteria for inclusion: (1) articles in English from peer-reviewed journals; (2) utilization of machine learning techniques; (3) focus on active military personnel, veterans, or both; (4) research aimed at evaluating suicide-related outcomes (e.g., suicidal ideation, suicide attempts, and suicide mortality); and (5) not a review article. 
A completed data analysis of the results was not a requirement for inclusion.

The focus of the study is solely on methods that use ML techniques. This work did not include traditional statistical approaches, such as pre-ML linear or logistic regression models, which have also provided valuable insights.

\subsection{Information sources}\label{information-sources}

The reviewers utilized several databases, including PubMed, Institute of Electrical and Electronics Engineers (IEEE) Xplore, Association for Computing Machinery (ACM) Digital Library, and Google Scholar. These databases were chosen for their relevance and comprehensive coverage of the scientific literature.

\subsection{Search strategy}\label{search-strategy}

A keyword literature search was performed in the databases on April 10th, 2024. With the keywords (``machine learning'') AND (``suicide'') AND (``veteran'' OR ``military'' OR ``recruit'' OR ``soldier''), there were 34 initial results in PubMed, 12 in IEEE, and 235 in ACM. Given a large number of initial hits on Google Scholar (N = 12,000) and the fact that many of these articles only mention these keywords in passing, a title search was conducted instead with the search term ``ti:(``machine learning'') AND (``suicide'') AND (``veteran'' OR ``military'' OR ``recruit'' OR ``soldier'')'' and the search returned 829 articles. We conducted another query using the keywords (``deep learning'') AND (``suicide'') AND (``veteran'' OR ``military'' OR ``recruit” OR “soldier'') on December 15th, 2024, in the same databases to include studies utilizing deep learning techniques. 

\subsection{Study selection}\label{study-selection}

During the two queries, after the initial exclusion and removal of duplicates, 35 articles proceeded to the phase of full-text review. Two reviewers (YZ and YW) conducted the full-text review. Three documents were excluded during this phase because they were either reviews or no specific machine learning model was adopted. A final set of 32 articles was found relevant and ready for data analysis (Figure~\ref{fig:flow}). Descriptive analyses were applied to analyze findings based on key design characteristics, such as study populations, specific machine learning techniques, and outcome measures.
The most statistically significant predictive factors for various suicidal outcomes were identified from articles and included in the tables as leading factors.

\section{Results}\label{result}

\subsection{Included studies and datasets}\label{included-studies-and-datasets}

Of the 1,110 articles identified during the initial keyword search, 32 met the inclusion criteria and were analyzed in this review. The articles covered predictions of multiple suicide outcomes, including thoughts of self-harm, suicidal ideation,  suicide attempts, suicidal ideation, and suicide death. The publication dates for the included papers range from 2014 to 2024. The findings of these articles were evaluated, and details regarding the study characteristics, such as sample size, data modality, and machine learning techniques used, are summarized in Table \ref{tab:overview}.


\begin{table}[t]
\centering
\caption{Overview of Included Studies. SI – Suicidal ideation. SA – Suicide attempts. SD – Suicide mortality, EMR– Electronic Medical Records, EHR– Electronic Health Records, HADS– Historical Administrative Data System.}
\label{tab:overview}
\tiny
\renewcommand{\tabularxcolumn}[1]{m{#1}}
\setlength{\tabcolsep}{1ex}
\rowcolors{2}{lightgray!30}{} 
\begin{tabularx}{\columnwidth}{llllrllX}
\toprule
\rowcolor[HTML]{EEECE1} 
Study & \makecell[l]{Outcome\\Measures} & Population & Country & \makecell[r]{ Sample\\ Size} & Data Sources & Metric(s) & Machine Learning Techniques \\
\midrule
\citet{gradus2017gender-s}  & SI & Veteran & US & 2,244 & Surveys & AUC: 0.91 & Random forest \\
\citet{colic2018using-z} & SI & Both & Canada & 738 & Surveys & AUC: 0.84 & Random forest \\
\cite{lin2020machine-n} & SI & Active duty & \makecell[l]{Taiwan,\\China} & 3,546 & Surveys & AUC: 1.00 & Logistic regression, decision   tree, random forest, support vector machine, multilayer perceptron \\
\citet{Belouali2021-ps} & SI & Veteran & US & 588 & \makecell[Xl]{Surveys,\\audio recordings} & AUC: 0.80 & Random forest, logistic regression, support vector machine, k-nearest neighbors, deep learning \\
\citet{borowski2022first-x} & SI & Veteran & US & 7,383 & Surveys & AUC: 0.89 & Random forest \\
\cite{vogt2024u-x} & SI & Veteran & US & 7,141 & Surveys & AUC: 0.83 & Cross-validated random forests \\
\citet{colic2022machine-v} & SI & Both & Canada & 738 & Surveys & AUC: 0.83 & Autoencoders, random forest, t-SNE (t-distributed Stochastic Neighbor Embedding) \\
\citet{zuromski2024detecting-c} & SI & Both & US & 8449 & Social media posts & F1: 0.73 & RoBERTa \\
\citet{thompson2014predicting-k} & SA & Veteran & US & - & EHR & Accuracy: 0.67 & MOSES \\
\citet{ben-ari2015text-q} & SA & Veteran & US & 250,401 & EHR & F1: 0.92 & Random forest \\
\citet{rosellini2017sexual-k} & SA & Active duty & US & 25,428 & HADS (STARRS) & - & Survival model \\
\citet{bernecker2019predicting-e} & SA & Active duty & US & 27,501 & HADS & AUC: 0.83 & Survival model, ensemble learning \\
\citet{zuromski2019assessment-q} & SA & Active duty & US & 3,649 & Surveys (STARRS) & AUC: 0.77 & Logistic regression \\
\citet{rozek2020using-r} & SA & Active duty & US & 152 & Surveys & \makecell[l]{Sensitivity: 0.30\\Specificity: 0.99} & ML methods based on agglomerative   supervised learning system \\
\citet{zuromski2020pre-deployment-z} & SA & Active duty & US & 7,677 & \makecell[l]{Surveys,\\HADS (STARRS)} & AUC: 0.78 & Penalized regression, elastic net   penalized regression, splines, random forest, regression trees \\
\citet{bryan2021improving-t} & SA & Both & US & 2,744 & Surveys & AUC: 0.82 & Logistic regression \\
\citet{mccarthy2021evaluation-v} & SA & Veteran & US & 173,313 & EHR & - & Regression \\
\citet{stanley2022predicting-j} & SA & Veteran & US & 8,899 & \makecell[Xl]{Surveys,\\HADS (STARRS)} & AUC: 0.74 & Isotonic regression, splines,   random forest, regression trees \\
\citet{chu2023associations-k} & SA & Veteran & US & 8,899 & Surveys (STARRS) & - & Statistical model \\
\citet{kearns2023practical-o} & SA & Veteran & US & 8,335 & Surveys  (STARRS) & AUC: 0.86 & Super Learner stacked generalization method \\
\citet{martinez2023deep-t} & SA & Veteran & US & $\sim$500,000 & EHR & PPV: 0.55 & Deep neural network \\
\citet{poulin2014predicting-g} & SM & Veteran & US & 210 & EHR & Accuracy: 0.65 & Supervised training with genetic   programming \\
\citet{kessler2015predicting-t} & SM & Active duty & US & 39,427 & HADS (STARRS) & AUC: 0.85 & Regression trees, elastic net \\
\citet{kessler2017predicting-s} & SM & Active duty & US & 148 & HADS (STARRS) & AUC: 0.75 & Naive Bayes, random forest,   support vector regression, elastic net penalized regression \\
\citet{kessler2017developing-c} & SM & Veteran & US & 27,480 & EMR & Sensitivity: 0.28 & Logistic regression, spline,   decision tree, support vector machines \\
\citet{kessler2020using-e} & SM & Veteran & US & 391,018 & \makecell[Xl]{EHR, \\admin data} & AUC: 0.79 & Logistic regression, decision   trees, regression trees, support vector machines, Neural networks \\
\citet{Lynch2020-hl} & SM & Veteran & US & 96,863 & EHR & - & Cox regression \\
\citet{levis2023leveraging-b} & SM & Veteran & US & 27,151 & EHR & AUC: 0.69 & Decision tree \\
\citet{Dhaubhadel2024-et} & SM & Veteran & US & 1,222,002 & EMR & C-index: 0.72 & Transfer learning \\
\citet{chakravarthula2020automatic-d} & SA, SI & Both & US & 124 & Audio    recordings & Recall: 0.60 & Support vector machine \\
\citet{brown2022digital-l} & SA, SI & Active duty & US & - & Surveys, biomarkers & - & Random forest \\
\citet{meerwijk2022suicide-d} & SA, SM & Veteran & US & - & EHR & - & Traditional NLP extractor and   embedding + regression\\
\bottomrule
\end{tabularx}
\end{table}

\subsection{Characteristics of datasets}\label{characteristics-of-datasets}

\paragraph{Data sources.}

The included studies utilized a variety of data sources. Twelve studies utilized surveys, questionnaires, audio recordings, and other self-reporting tools to collect data on factors like suicidal ideation, substance misuse, PTSD symptoms, overall well-being, and major life events. These studies included both single-center and multicenter settings, with some studies recruiting participants from a single institution and others from national samples. Reporting the level of care was not common, but there were exceptions. For example, Bryan et al. specified that their research was conducted in a primary care environment.\cite{bryan2021improving-t} Ten studies relied on Electronic Health Records (EHR) systems, including those from the Veterans Health Administration (VHA)'s Corporate Data Warehouse (CDW). Additionally, eight used data from the Army Study to Assess Risk and Resilience in Service-members (Army STARRS); of these, four accessed administrative data from the Historical Administrative Data System (HADS).

The use of EHR, particularly natural language processing (NLP) of free-style clinical notes, has expanded considerably in medical research. Among the studies we reviewed, five utilized textual data from these records, and one adopted a unique approach by using audio recordings of conversations between military couples as their data source.\cite{chakravarthula2020automatic-d}
One study also used social media posts.\cite{zuromski2024detecting-c} 
The remaining studies took a multimodal approach, often combining elements such as surveys with data from Army STARRS.

\paragraph{Sample sizes.}

The sample sizes in the studies reviewed ranged from 124 to 1,222,002 (Mean = 98,353, SD = 248,044, Median = 8,335). Although most studies utilized large datasets, several ($n = 3$) relied on significantly smaller sample sizes, with only one or two hundred participants. This could limit the generalizability of those specific findings.

\begin{figure}
\centering
\includegraphics[width=\textwidth]{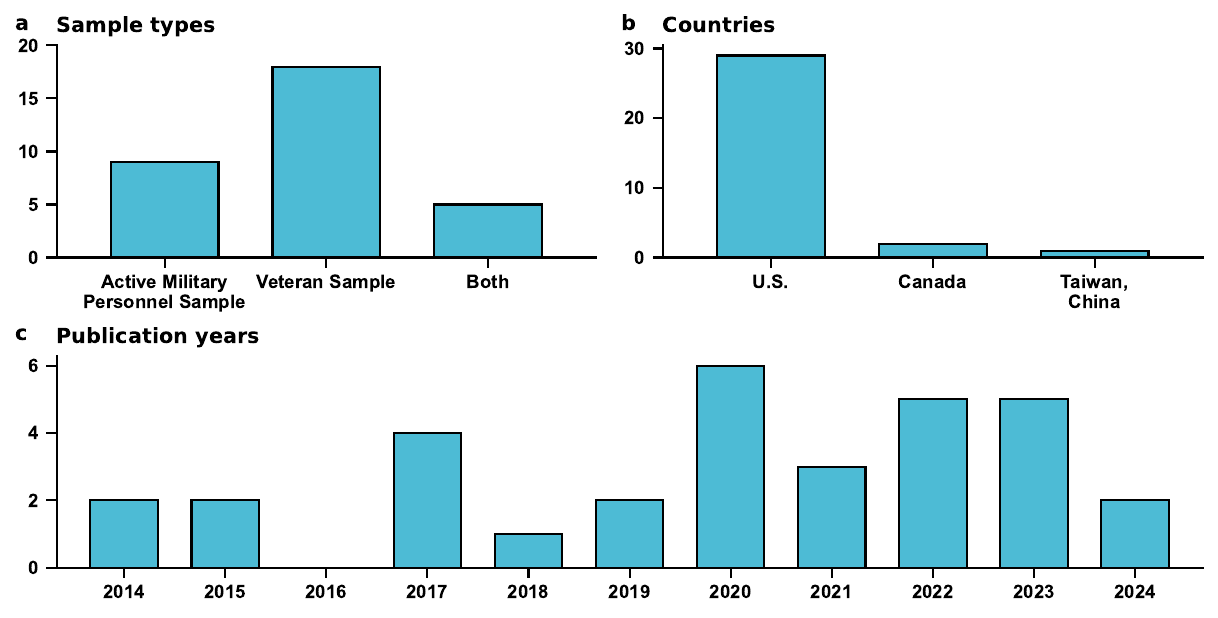}
\caption{Characteristics of included studies. (A) Sample types. (B) Countries. (C) Publication years.}
\label{fig:characteristics}
\end{figure}


\paragraph{Sample characteristics.}

Nine exclusively studied current military personnel, eighteen focused solely on veterans, and five assessed both active service members and veterans (Figure \ref{fig:characteristics}). Geographically, most of the studies were conducted in the U.S. ($n = 29$), with the remainder conducted in Canada ($n = 2$)\cite{colic2018using-z, colic2022machine-v} and Taiwan, China ($n = 1$).\cite{lin2020machine-n} A majority of studies ($n = 20$) did not report the racial composition of their samples. Among those that did, Caucasians were the most represented group, with their participation ranging from 59.13\% to 82\%. This representation significantly exceeds that of other racial groups in the samples.

Among nineteen studies that specified sex composition, the proportion of male participants ranged from 68.4\% to 100\%. This reflects the predominantly male makeup of the military.\cite{Parker_undated-du}
However, three studies focused on specific subjects like couples' conversations or sex differences, resulting in almost equal numbers of male and female participants in their samples. Additionally, one study specifically targeted female soldiers who had been sexually assaulted and therefore included no male participants.\cite{rosellini2017sexual-k}

\paragraph{Clinical conditions.} 

Several studies selected participants based on specific health-related criteria. Specifically, Belouali et al. focused on individuals with Gulf War Illness.\cite{Belouali2021-ps}
At the same time, Rozek et al. analyzed participants with active suicidal ideation and a history of suicide attempts,\cite{rozek2020using-r} and Rosellini et al. concentrated on those with a history of sexual assault.\cite{rosellini2017sexual-k}
Additionally, Gradus et al. specifically included participants with suicidal ideation with symptoms associated with mental health conditions like probable depression, anxiety, and PTSD.\cite{gradus2017gender-s}
Moreover, seven studies examined individuals with a history of health services usage, such as outpatient visits or inpatient hospitalizations. Among these studies, Lynch et al. specifically targeted sexual minorities with a history of VHA outpatient service use.\cite{Lynch2020-hl}

\paragraph{Outcome variables.}

The primary focus of most studies was limited to the prediction of either suicidal ideation (n = 8), suicide attempt ($n = 13$), or suicide mortality ($n = 8$). Additionally, three studies aimed to assess/predict more than one of these outcomes. In addition, one study examined the prediction of self-harm thoughts along with suicidal ideation (SHSI).\cite{colic2022machine-v}
We also observed substantial variability in the time frames for suicide prediction across the studies, ranging from immediate current status assessments ($n=1$) to long-term forecasts extending up to 10 years ($n=24$). Specifically, three studies focused on predictions within six months, eleven within twelve months, and ten studies aimed at predicting outcomes for periods longer than one year.

\subsection{Machine learning methodologies}\label{machine-learning-methodologies}

\subsubsection{Methods}

The choice of machine learning techniques in the reviewed studies is influenced by factors, such as sample size, data modality, and research objectives. The majority of studies ($n = 21$) employed traditional machine learning or statistical models, typically in small to medium-sized samples (fewer than 10,000 participants). We identified only three studies that used deep learning-based methods.\cite{martinez2023deep-t}

As discussed above, survey data, predominantly consisting of categorical and numerical variables, was the most common data modality. Given the typical limitation in the number of variables, traditional statistical learning methods like logistic regression, random forests, and support vector machines (SVM) were frequently utilized. These methods can handle datasets with a manageable number of features and are highly interpretable.

Additionally, several studies implemented ensemble techniques to enhance the accuracy of their predictions.\cite{Dhaubhadel2024-et, bernecker2019predicting-e}
This approach combines the predictions from multiple machine learning models to improve overall predictive performance. It helps mitigate the weaknesses of individual models by leveraging their collective strength, leading to more reliable and precise outcome predictions.

To process free-style texts, most papers employed traditional NLP methods. For instance, \citet{levis2023leveraging-b} implemented the Term Frequency-Inverse Document Frequency (TFIDF) technique for feature extraction, while Poulin et al. used the bag-of-words approach.\cite{poulin2014predicting-g}
Additionally, Chakravarthula et al. extracted traditional speech recognition features, including pitch, intensity, voice quality, and Mel Frequency Cepstral Coefficients (MFCC) from audio recordings.\cite{chakravarthula2020automatic-d}

\subsubsection{Prediction performance and metrics}

The area under the receiver operating characteristic curve (AUC) was the most widely used metric of predictive accuracy in the reviewed studies, with most showing values around 0.80 to 0.85. The specific AUC value depended on both the model of choice and the underlying dataset. Besides AUC, some studies used metrics such as PPV (positive predictive value), NPV (negative predictive value), sensitivity, and specificity. Although several studies frame the ML problem as a survival model,\cite{rosellini2017sexual-k, bernecker2019predicting-e} none used survival-specific metrics such as c-index. In cases where the focus was on the application of machine learning rather than the methodology, performance metrics were often not reported.



\subsection{Predominant risk factors for suicidal behavior identified by ML}\label{important-risk-factors}

The included studies highlighted a diverse set of risk factors crucial for forecasting suicide-related outcomes. We summarized the key factors and categorized them according to the guidelines outlined in the National Violent Death Reporting System Web Coding Manual  (Table \ref{tab:risk description} and Table \ref{tab:risk}).\cite{Other2022national-f} Our analysis also examined sex-specific factors (Table \ref{tab:gender}).


\begin{table}
\centering
\caption{Description of the types of risk factors associated with suicide as identified in the included articles.}
\label{tab:risk description}
\footnotesize
\rowcolors{2}{lightgray!30}{} 
\newcolumntype{s}{>{\hsize=.25\textwidth\raggedright\arraybackslash}X}
\begin{tabularx}{\textwidth}{sX}
\toprule
\rowcolor[HTML]{EEECE1}
Factors & Description \\
\midrule
Alcohol abuse & Any factors related to alcohol dependence/alcohol abuse   problem \\
Childhood adversity & Adverse experiences during childhood, such as parent   maltreatment and impaired parenting ability \\
Criminal legal problem & Criminal legal problems that are relevant to suicide \\
Demographics & Sex, age, race, ethnicity, marital status, relationship   status, sexual orientation \\
Factors from psychological assessment & Specific items from psychological assessments used (e.g.,   PHQ – 2f, “Feeling bad/like a failure/ let people down”)\\
Financial problem & Any financial difficulties that serve as suicide   predictors \\
History of mental illness treatment & History of ever being treated for a mental health problem \\
History of self-harm & Factors related to having a history of intentionally   inflicting pain on one’s own body without the conscious intent of dying by   suicide \\
History of substance abuse treatment & History of ever being treated for substance/alcohol abuse   problem \\
Interpersonal violence-perpetrator & Being the perpetrator of any type of interpersonal   violence \\
Interpersonal violence-victim & Victimization of any type of interpersonal violence   (physical, verbal, sexual) \\
Involvement with the criminal justice system & Any involvement with the criminal justice system.\\
Job problem & Factors related to general job problems that contribute to   suicide \\
Lifetime stressful/traumatic event & Any significant life event that causes distress or traumatic experiences (e.g. being bullied by the unit/spending time in jail) \\
Mental health problem & Any current or lifetime mental health illness (e.g.,   depression, anxiety, PTSD) \\
Military experience-relevant factor & Any factor that is related to military experience, such as   prior deployment, unit cohesion, bully in unit, enlisted rank, warfare   exposure \\
Other healthcare resources utilization history & Factors relevant to usage of healthcare resources other than mental health/substance abuse treatments (e.g., emergency department visits) \\
Other psychological factors & Any psychological predictors except mental health problems   (e.g., psychological resilience) \\
Other substance abuse & Any non-alcohol related substance abuse problem. \\
Personality & Factors related to specific personalities (e.g., negative affect, fearlessness) \\
Physical health problem & Physical health problems that may affect life quality and   are relevant to suicide \\
Relationship problem & Problems with people around that contribute to suicide   (e.g., lack of social support, feelings of being a burden, low relationship quality) \\
Suicide attempt history & Factors related to having a history of attempting suicide \\
Suicide thought history & Factors related to having a history of suicidal thoughts or plans. Disclosure can be verbal, written, or electronic. \\
Traumatic brain injury & Any factor related to traumatic brain injury such as   lifetime experience, worst severity, and post-concussive symptoms \\
Weapon problem & Suicide predictors relevant to weapons (e.g., problematic   weapon possession) \\
\bottomrule
\end{tabularx}
\end{table}

\begin{table}
\centering
\caption{Important factors influencing suicidal ideation, attempt, and mortality in specific sub-populations.}
\label{tab:risk}
\footnotesize
\begin{tabularx}{\textwidth}{l
>{\hsize=0.7\hsize\raggedright\arraybackslash}X
>{\hsize=1.3\hsize\raggedright\arraybackslash}X}
\toprule
\rowcolor[HTML]{EEECE1} 
Outcome & Population & Leading factors \\
\midrule
Suicidal ideation & General veteran   population & Mental health problems,   other psychological factors, military experience-related factors, physical   problem, relationship problem, financial problem, alcohol abuse, traumatic   brain injury \cite{vogt2024u-x, borowski2022first-x}\\
\cmidrule{2-3}
 & Veterans with suicidal   ideation and mental health problems & Mental health problem,   physical health problem, military experience-relevant factors, interpersonal   violence-victim \cite{gradus2017gender-s}\\
\cmidrule{2-3}
 & Both current service   members and veterans & Factors from   psychological assessment \cite{colic2018using-z}\\
\midrule
Suicide attempt & General active service   members population & Childhood adversity,   lifetime stressful/traumatic event, particular personality, mental health   problem, suicide thought history, relationship problem, physical health   problem, traumatic brain injury, history of mental illness treatment,   demographic factors, military experience-relevant factors \cite{zuromski2020pre-deployment-z}\\
\cmidrule{2-3}
 & Current service members   who deny suicidal ideation & Mental health problem,   lifetime stressful/traumatic event \cite{bernecker2019predicting-e}\\
\cmidrule{2-3}
 & Current service members   with current suicidal ideation and/or attempt & Suicide thought   history, suicide attempt history, mental health problem, military   experience-related factor, demographics, history of mental illness treatment \cite{rozek2020using-r, zuromski2019assessment-q}\\
\cmidrule{2-3}
 & General veteran   population & Suicide thought   history, lifetime stressful/traumatic events, involvement with the criminal justice system, financial problem, physical problem, suicide attempt history, demographics, other   healthcare resources utilization history, alcohol abuse, mental health   problem, history of mental illness treatment, job problem, financial problem \cite{martinez2023deep-t, kearns2023practical-o, stanley2022predicting-j, ben-ari2015text-q, chu2023associations-k}\\
\midrule
Suicide mortality & Current service members   after psychiatric outpatient visits & Involvement with the criminal justice system,   mental health problem, other healthcare resource utilization history,   demographics, military experience-related factors, interpersonal   violence-perpetrator, weapon problem, suicide attempt history, history of   mental illness treatment, demographics, interpersonal violence, military   experience-related factor \cite{kessler2017predicting-s}\\
\cmidrule{2-3}
\cmidrule{2-3}
 & Veterans with   psychiatric hospitalizations & Suicide attempt   history, history of mental illness treatment, substance abuse other,   demographics \cite{kessler2015predicting-t, kessler2020using-e}\\
\bottomrule
\end{tabularx}
\end{table}

\begin{table}
\centering
\caption{Sex-specific factors associated with suicidal ideation only.}
\label{tab:gender}
\footnotesize
\begin{tabularx}{\textwidth}{
>{\hsize=0.7\hsize\raggedright\arraybackslash}X
l
>{\hsize=1.3\hsize\raggedright\arraybackslash}X}
\toprule
\rowcolor[HTML]{EEECE1} 
Population & Sex & Leading factors \\
\midrule
General veterans’ population & Male & Mental health problem,   other psychological factor, relationship problem, physical health problem,   financial problem, military experience-related factor, traumatic brain injury \cite{borowski2022first-x}\\
\cmidrule{2-3}
 & Female & Mental health problem,   physical health problem, relationship problem, other psychological factor,   military experience-related factor, demographics, financial problem, alcohol   abuse \cite{borowski2022first-x}\\
\midrule
Veterans with suicidal   ideation and mental health problems & Male & Mental health problem,   physical health problem, military experience-related factor, lifetime   stressful/traumatic event, demographics, alcohol abuse \cite{gradus2017gender-s}\\
\cmidrule{2-3}
 & Female & Mental health problem,   interpersonal violence-victim, alcohol abuse \cite{gradus2017gender-s}\\
\bottomrule
\end{tabularx}
\end{table}



\subsubsection{Factors associated with suicidal ideation}\label{factors-associated-with-suicide-ideation}

In the general veteran population, self-reported depression symptom severity was identified as an important predictor for males. Other key factors for this group include the severity of other mental health disorders (e.g., anxiety and PTSD), self-identified depression, and overall health and well-being.\cite{borowski2022first-x}
For female veterans, similar factors were found, with depression severity also being a leading factor. However, unlike for male veterans, self-identified anxiety, age, and military service duration were less statistically significant for female veterans.\cite{borowski2022first-x}
Additional leading factors specific to female veterans included pay grade, self-reported chronic pain, alcohol misuse severity, and lack of social involvement. PTSD ranked as the 16th most important factor of suicidal ideation out of 123 factors for both male and female veterans.\cite{borowski2022first-x}

Further analysis among veterans with suicidal thoughts and mental health conditions also revealed distinct sex-specific factors. For males in this group, the key factors included probable depression (symptoms were assessed using standardized scales rather than through formal diagnoses), anxiety, and deployment experience in Iraq or Afghanistan. For females in this group, while probable depression was still an important predictor, other major factors were experiences of sexual harassment during deployment and probable PTSD.\cite{gradus2017gender-s}

\subsubsection{Factors associated with suicide attempts}\label{factors-associated-with-suicide-attempts}

Among service members, several similar factors emerged as predictive of suicide attempts, such as childhood adversities, traumatic events throughout life, personality traits, and history of mental disorders. Demographic factors and military-related experiences, such as age, sex, education, marital status, previous deployments, and unit cohesion, also emerged as factors. 
Additionally, number of months having experienced PTSD in the last year was noted as being a statistically significant factor associated with suicide attempts.\cite{zuromski2020pre-deployment-z} 
Bernecker et al. found that for soldiers who did not acknowledge suicidal thoughts, certain factors were still important in predicting suicide attempts.\cite{bernecker2019predicting-e}
These factors included severity scores for depression, PTSD (especially a combination of lifetime PTSD severity and other mental disorders such as depression, anxiety, and irritability), experiences of being bullied in their unit, and time spent in jail.

For military personnel who attempted suicide, key predictive factors included the severity and nature of suicidal experiences (e.g., the intensity of suicidal thoughts at their worst, suicidogenic cognitions like hopelessness), military-related factors (e.g., enlisted rank), and demographic factors (e.g., male). Additionally, whether they received specific treatments, such as Brief Cognitive Therapy, also helped to predict suicide attempts.\cite{rozek2020using-r, zuromski2019assessment-q}
In this particular group, PTSD was not a major factor in predicting suicide attempts and mortality, despite many in the sample having a PTSD diagnosis.\cite{rozek2020using-r, kessler2015predicting-t}

In the veteran population, Stanley et al. reported that the most effective factor was lifetime suicide plans, followed by lifetime trauma, poverty level, and property crimes committed within the previous 12 months.\cite{stanley2022predicting-j}
Other effective factors were age, visits to the emergency department, alcohol dependence, and depressive disorder.\cite{martinez2023deep-t, ben-ari2015text-q, chu2023associations-k}
It is noteworthy that there were mixed findings regarding the importance of mental health disorders in predicting suicide attempts for veterans. Some studies highlight key mental health-relevant factors, while others suggest that few are among the leading factors.\cite{stanley2022predicting-j}

\subsubsection{Factors associated with suicide death}\label{factors-associated-with-suicide-mortality}

Among soldiers who had received psychiatric outpatient care, committing crimes was an important factor. Moreover, mental and physical health factors continued to be statistically significant. Among soldiers with psychiatric hospitalizations, other effective predictors included being male, enlisting at an older age, exhibiting verbal violence towards others, and having a history of psychiatric treatment.\cite{kessler2015predicting-t}

For veterans with psychiatric hospitalizations, past suicide attempts, history of psychiatric hospitalizations, and being non-Hispanic African American race were all considered important. On the other hand, medications classified to increase suicide risks and medical procedures had minimum effect.\cite{kessler2020using-e}

\subsection{Mitigation and intervention}

While the primary application of ML in these reports was to identify suicide-related factors, some studies provided valuable insights into how to mitigate these factors. In this review, we observed two intervention strategies. The first involved consistent, brief interventions for a larger population with relatively low-risk scores. The second focused on intensive interventions for a smaller group with high-risk scores. The former strategy required the model to have high recall, whereas the latter emphasized precision. Although some studies suggested that brief and consistent interventions for large populations could be more beneficial42, most interventions in practice target smaller populations.\cite{borowski2022first-x} This preference is often due to the side effects of interventions, which can include negatively impacting a person's career such as early termination of military service.\cite{zuromski2019assessment-q, kessler2015predicting-t}

Compared with other screening approaches, ML methods have unique advantages. First, ML models may quickly identify previously unnoticed risk factors. For instance, social determinants of health, which are often overlooked, have been identified as important.\cite{kessler2020using-e} 
Second, ML models can predict various time horizons, allowing for different intervention strategies.\cite{bernecker2019predicting-e} Lastly, as more health records become available, ML may adapt the risk score in real time, offering continuous monitoring. This capability is particularly valuable for providing interventions during the critical period when risk scores spike suddenly.\cite{thompson2014predicting-k}

\section{Discussion}\label{discussion}

Suicide is a significant issue among service members and veterans, and machine learning is uniquely positioned to enhance our understanding of this issue. It offers the potential to identify at-risk individuals early, thus providing vital lead time for implementing preventative measures and treatment interventions. This proactive approach could significantly enhance the effectiveness of suicide prevention strategies.

Compared to traditional approaches, we find that ML methods can handle large datasets, accommodate diverse data modalities, and achieve higher prediction accuracy. They can automatically capture complex relationships among covariates to predict suicide outcomes from vast amounts of data. While predicting suicide remains a challenging problem with significant opportunities for ML models, these models have already shown noticeable improvements over traditional statistical methods in most studies, including improved accuracy in predicting suicidal ideation, attempted suicides, and suicide mortality. 
In addition, increasing evidence indicates that these models may also excel in predicting self-harm outcomes.\cite{colic2022machine-v}
Their ability to analyze vast datasets quickly and identify patterns that may not be immediately obvious to human observers makes them a potentially invaluable tool in the ongoing fight against suicide and self-harm among military personnel and veterans. 

The studies reviewed in this manuscript demonstrate that risk factors associated with suicide in military contexts span a wide range of psychological, biological, and sociocultural factors, in accordance with many findings in previous non-ML studies. Sex-based differences were also identified. These findings highlight the challenges involved in assessing suicide risk among service members and veterans. 
This complexity underscores the need for holistic, personalized support systems and intervention strategies. Collaboration among behavioral health professionals, community workers, and machine learning experts will be highly beneficial. The diversity of these factors, which include aspects such as mental health status, personal history, environmental influences, and situational factors, demonstrates that effective prevention strategies must be comprehensive and flexible. They should be designed to address each individual's unique circumstances and needs, ensuring that interventions are relevant and responsive to the diverse challenges faced by service members and veterans.

When comparing risk factors between male and female veterans, most factors were found to be similar. However, alcohol misuse and sexual abuse were more statistically significant factors among females, while traumatic brain injury (TBI) emerged as a more prominent issue among males. Previous research has highlighted binge drinking as a particular issue in the female veteran population. 
However, female veterans generally consume less alcohol than male veterans. 
They can experience similar levels of alcohol-related problems at lower consumption levels due to lower tolerance thresholds.\cite{Bradley2001-tj} 
Findings also suggest a more rapid transition from recreational to clinically significant substance use among females versus males.\cite{towers2023sex/gender-h}
The heightened impact of TBI in male veterans can be attributed to higher incidence rates of TBI in this group, coupled with potentially greater stigma and barriers to seeking mental health support.\cite{Bahraini2013-zn}

While PTSD and other mental health conditions are widely recognized as significant factors associated with suicide in military-connected populations,\cite{pompili2013posttraumatic-q, stanley2022predicting-j} many studies in this literature review did not delve deeply into how they specifically contribute to suicide-related outcomes or its underlying psychological mechanisms. Moreover, the role of PTSD and other mental health conditions varies considerably across different study samples. For instance, it is found to be a crucial factor in general samples of active service members and veterans, aligning with past research.\cite{jakupcak2009posttraumatic-p}
However, its significance diminishes in specific groups, such as those exhibiting active suicidal ideation.\cite{rozek2020using-r}

Lastly, service members/veterans share some risk factors with civilians. Research indicates that childhood adversities, such as sexual abuse, physical punishment, and financial hardship before the age of 7, significantly impact members of the general population.\cite{hardt2008childhood-t} Traumatic events occurring before age 18 have also been identified as a single strong predictor of suicide and suicide attempts.\cite{gunter2011predictors-f} 
Personality traits, including negative affectivity and impulsivity, also play a significant role in predicting suicide attempts.\cite{yen2009personality-g} 
Mental illnesses, including depression and anxiety, can also predict suicide risk in all these populations.\cite{ehtemam2024role-k} 
Similarly, certain demographic factors such as age and sex are considered common risk factors as well.\cite{ehtemam2024role-k} 
Alcohol and substance abuse also predict suicide outcomes in the veteran population and in the general public,\cite{borowski2022first-x, ehtemam2024role-k} even though these factors are not as strong in predicting military personnel's suicide risks, possibly due to military lifestyle and regulations. However, veterans and current service members face unique military-relevant risk factors, including combat exposure, enlisted ranks, and having to endure the great changes in life due to transitioning. Additionally, marital status is a weaker predictor in military personnel/veterans compared to the general public. 

The current research on ML-based prediction of suicide among military personnel and veterans has several important limitations that future studies need to address.

\paragraph{Limited scope of data sources:} Currently, the primary data sources are surveys and Electronic Health Records (EHRs). Incorporating additional and diverse data types, such as online social media content and data from wearable technology, could potentially enhance the accuracy of predictive models. For example, considering how prevailing models of suicide attempts attribute these to an intersection of sociodemographic, psychiatric, or other trait risk factors (typically identified by ML analyses) with individual cognitive or dispositional factors\cite{oConnor2014psychology-k}, low-burden cognitive assessments could be blended with administrative data, where ML could be leveraged to identify particularly problematic interactions of multiple factors. 
For studies included in this review, Thompson et al. represented a pioneering effort, utilizing social media data for assessing suicide risk.\cite{thompson2014predicting-k}
Given the structured nature of military environments, obtaining consent from military personnel for using such data in research might be more feasible. Moreover, since wearable devices are commonly utilized among service members, they could serve as valuable data sources for continuous monitoring. Expanding the variety of data sources not only enriches the dataset but also allows for the full capabilities of machine learning and artificial intelligence techniques to be employed more effectively. 
Wearable technology, in particular, can provide real-time, continuous data that reflects physiological and behavioral patterns, such as heart rate and daily activity levels. In addition, smartphone apps can now passively capture the semantic content of texts and other verbalizations that are showing evidence of predicting or serving as markers of depression or suicidality.\cite{efe2024linguistic-y} This capability offers potential insights into stress levels, sleep quality, and overall mental health status that are not typically captured through traditional surveys or EHRs.\cite{brown2022digital-l}

\paragraph{Methodological enhancements:} Due to the highly organized nature of the military system, there exist extensive datasets with large sample sizes, including surveys and EHRs. However, the ML methodologies employed to date predominantly originate from the pre-deep learning era. Employing more advanced deep learning and machine learning techniques might enhance the predictive accuracy and reliability of the studies, providing stronger evidence for the identified risk factors and mitigation strategies. For instance, despite the availability of extensive clinical notes from EHR and some conversational datasets, they are mainly processed using traditional NLP methods. There is a substantial opportunity to adopt more advanced technologies, such as transformers, which could better harness the informational potential of these extensive data sources.

Although several studies include longitudinal features, they are often treated as scalar variables. Risk factors identified through longitudinal models are generally more plausible and have a higher potential for establishing causality. Collecting more longitudinal data and adapting machine learning techniques are other important directions for methodology advancements.

\paragraph{Better ML workflow:} Most studies followed standard ML workflows, including the separation of training, validation, and test datasets, as well as regularization techniques to prevent overfitting. However, several research gaps could be improved in future research. First, several studies overlooked metrics that differentiate false positives and false negatives, such as PPV (precision), NPV, or sensitivity. In the context of suicide, these distinctions are important, and a more detailed error analysis is needed. Second, while survival analysis provides an appropriate framework for the problem, survival-specific metrics such as the c-index should be used. Third, one study reported an AUC of 1.00, which needs further clarification and explanation. Lastly, most studies focus solely on the ML aspect without discussing its connection to clinical rationales.

\paragraph{Diversity and cultural representation:} Most studies primarily focus on U.S. samples, leading to an underrepresentation of service members and veterans from other countries and regions. Expanding research to include these groups would enhance the generalizability of the findings and provide insights into the global aspects of military mental health. Given that the percentage of participants identified as White race in the analyzed samples was higher than that of other racial groups, there might also be a need for more focused research on racial minority groups in this field. Performing cross-cultural comparisons could uncover unique factors of suicide and mental health patterns that vary by culture and ethnicity, thereby enhancing the applicability and effectiveness of intervention strategies.

\paragraph{Dedicated analyses on specific subgroups:} The military community could benefit from more detailed analyses targeting specific subgroups, particularly those with certain mental or physical conditions, such as individuals with active suicidal ideation or those frequently utilizing mental health outpatient services. Such focused studies could improve external validity and provide deeper insights into these groups' distinct needs and risk factors, ultimately providing better support for these vulnerable populations. In addition, although findings suggest some sex-based differences, additional efforts are required to ensure that sufficient sample sizes of females are included to ensure the identification of risk factors pertinent to females who have served.

\paragraph{More focus on PTSD:} The importance of PTSD increases when predicting self-harm thoughts and suicidal ideation (SHSI) compared to suicidal ideation alone, within the same mixed population of both active service members and veterans.\cite{colic2018using-z, colic2022machine-v}
Despite the extensive research performed on how PTSD contributes to suicide-related outcomes, there is a notable gap in research specifically analyzing PTSD as a risk factor in SHSI. Additionally, there is limited understanding of why the effects of PTSD diminish when self-harm is excluded from the prediction.

\paragraph{Ethical considerations:} Although most studies did not address ethical considerations, acknowledging their importance is essential for future considerations. One key ethical concern is bias. Some ML models predicting suicide-relevant outcomes may potentially discriminate against racial groups such as African Americans/Native Americans, since they are underrepresented in the training data. This bias could result in more false positives, leading to unnecessary outreach and/or treatments. Another concern is transparency. While many studies have advocated for model sharing, it is evident that simply sharing algorithms without the associated data does not ensure transparency. To enable external validation of algorithms, both the data and underlying code should be shared whenever possible.\cite{kirtley2022translating-j}

\section{Conclusion}

Military personnel and veterans are particularly vulnerable to suicide-related outcomes. However, current methodologies to identify and manage suicide-related risk factors remain limited. With the increasing volume of structured data in the military/veteran community, machine learning methods offer the potential for more effective, data-driven decision-making. In this work, we reviewed the application of machine learning methods in this domain, summarized the risk factors identified by these models, and suggested future research directions, including incorporating more diverse data sources, employing more advanced modeling techniques, focusing on specific groups, and paying more attention to the role of PTSD in SHSI. We hope this review will increase awareness and contribute to advancing the ML techniques in this critical area.

\paragraph{Funding statement.}\label{acknowledgment}

This study is supported by funding from the Artificial Intelligence/Machine Learning Consortium to Advance Health Equity and Researcher Diversity (AIM-AHEAD), a program of the National Institutes of Health (1OT2OD032581-02-259; Dr Peng). This presentation is based on work supported, in part, by the Department of Veterans Affairs (VA), but does not necessarily represent the views of the VA or the United States Government.

\paragraph{Competing interests statement.}\label{competing-interests}

The Authors declare no Competing Financial or Non-Financial Interests.

\paragraph{Data availability statement.}

The data underlying this article are available in the article.

\paragraph{Contributorship statement}\label{author-contributions}

Study concepts/study design, Y.Z., Y.W., Y.P.; manuscript drafting or manuscript revision for important intellectual content, Y.Z., Y.W., Y.W., R.K.P., G.L.H., Y.Z., C.W., J.L., L.A.B., J.M.B., Y.P; approval of the final version of the submitted manuscript, all authors; agrees to ensure any questions related to the work are appropriately resolved, all authors; literature research, Y.Z., Y.W.; experimental studies, G.H., M.L.; data interpretation and statistical analysis, Y.Z., Y.W., L.A.B., J.B.; and manuscript editing, all authors.

\bibliographystyle{medlinenat}
\bibliography{main}




\end{document}